\pdfoutput=1

\documentclass[11pt]{article}

\usepackage{acl}

\usepackage{times}
\usepackage{latexsym}

\usepackage[T1]{fontenc}

\usepackage[utf8]{inputenc}

\usepackage{microtype}

\usepackage{inconsolata}

\usepackage{soul}
\usepackage{amsmath}
\usepackage{booktabs}
\usepackage{bm}
\usepackage{cleveref}
\usepackage{graphicx}
\usepackage{tabularx}
\usepackage{soul}
\usepackage{xcolor}
\usepackage{todonotes}
\usepackage{multirow}
\usepackage{xpatch}
\usepackage{blindtext}
\usepackage{fdsymbol}
\usepackage{microtype}
\usepackage{hyperref}
\usepackage{adjustbox}
\usepackage{textcomp}
\usepackage{xparse}
\usepackage{enumitem}
\usepackage{makecell}
\usepackage{subcaption}
\usepackage{colortbl}

\usepackage{pifont}

\usepackage{xparse}
\usepackage{stfloats}

\crefformat{section}{\S#2#1#3} %
\crefformat{subsection}{\S#2#1#3}
\crefformat{subsubsection}{\S#2#1#3}

\newcommand{\samsum}[1]{\textsc{SAMSum}}
\newcommand{\dialogsum}[1]{\textsc{DialogSum}}
\newcommand{\mixandmatch}[1]{\textsc{MixAndMatch}}
\newcommand{\confit}[1]{\textsc{ConFiT}}
\newcommand{\ctrldiasumm}[1]{\textsc{CtrlDiaSumm}}
\newcommand{\cods}[1]{\textsc{CODS}}
\newcommand{\modelshort}[1]{\textsc{ZeroFEC}}
\newcommand{\modelshortda}[1]{\textsc{ZeroFEC-DA}}
\newcommand{\datashort}[1]{\textsc{ManiTweet}}
\newcommand{\politifact}[1]{\textsc{PolitiFact}}
\newcommand{\gossipcop}[1]{\textsc{GossipCop}}
\newcommand{\fakenewsnet}[1]{\textsc{FakeNewsNet}}
\newcommand{\recontextualized}[1]{\textsc{Mani}}
\newcommand{\notrecontextualized}[1]{\textsc{NoMani}}

\definecolor{c2}{RGB}{218,0,0}

\definecolor{lightblue}{RGB}{212, 235, 255}
\definecolor{lightorange}{RGB}{255, 204, 168}
\definecolor{lightyellow}{RGB}{255, 255, 168}
\definecolor{lightred}{RGB}{255, 168, 168}
\definecolor{darkred}{RGB}{196, 30, 58}
\definecolor{lightgreen}{rgb}{0.82, 0.94, 0.75}
\definecolor{darkgreen}{rgb}{0.56, 0.63, 0.51}
\definecolor{lightgray}{rgb}{0.7, 0.7, 0.7}
\definecolor{gold}{rgb}{0.83, 0.69, 0.22}
\sethlcolor{lightblue}
\newcolumntype{Y}{>{\centering\arraybackslash}X}
\newcommand\hlc[2]{\sethlcolor{#1} \hl{#2}}

\NewDocumentCommand{\steeve}
{ mO{} }{\textcolor{gold}{\textsuperscript{\textit{Steeve}}\textsf{\textbf{\small[#1]}}}}
\NewDocumentCommand{\heng}
{ mO{} }{\textcolor{red}{\textsuperscript{\textit{Heng}}\textsf{\textbf{\small[#1]}}}}

\NewDocumentCommand{\ken}
{ mO{} }{\textcolor{blue}{\textsuperscript{\textit{Ken}}\textsf{\textbf{\small[#1]}}}}

\definecolor{gold}{rgb}{0.83, 0.69, 0.22}

\newcommand{\cmark}{\ding{51}}%
\newcommand{\xmark}{\ding{55}}%

\title{\datashort~: A New Benchmark for Identifying \\Manipulation of News on Social Media}%

\author{Kung-Hsiang Huang$^{1,3*}$ ~~~Hou Pong Chan$^{1}$ ~~~ Kathleen McKeown$^{2}$ ~~~ Heng Ji$^{1}$\\
$^{1}$University of Illinois Urbana-Champaign ~~~ 
$^{2}$Columbia University ~~~ $^{3}$Salesforce AI Research \\
$^{1}$\texttt{\{hpchan, hengji\}@illinois.edu} 
$^{2}$\texttt{kathy@cs.columbia.edu} ~~~ $^{3}$\texttt{kh.huang@salesforce.com}
\\}
\begin{document}
\maketitle
{\def\thefootnote{*}\footnotetext{Work was done while Kung-Hsiang was at UIUC.}}
\begin{abstract}

Considerable advancements have been made to tackle the misrepresentation of information derived from reference articles in the domains of fact-checking and faithful summarization. However, an unaddressed aspect remains - the identification of social media posts that manipulate information presented within associated news articles. This task presents a significant challenge, primarily due to the prevalence of personal opinions in such posts. We present a novel task, \textit{identifying manipulation of news on social media}, 
which aims to detect manipulation in social media posts. %
To study this task, we have proposed a data collection schema and curated a dataset called \datashort~, consisting of 3.6K pairs of tweets and corresponding articles. Our analysis demonstrates that this task is highly challenging, with large language models (LLMs) yielding unsatisfactory performance. Additionally, we have developed a simple yet effective framework that outperforms LLMs significantly on the \datashort~ dataset. Finally, we have conducted an exploratory analysis of human-written tweets, unveiling intriguing connections between 
manipulation and factuality of news articles. \looseness=-1%

\end{abstract}

\begin{figure}[bt]
    \centering
    \includegraphics[width=0.9\linewidth]{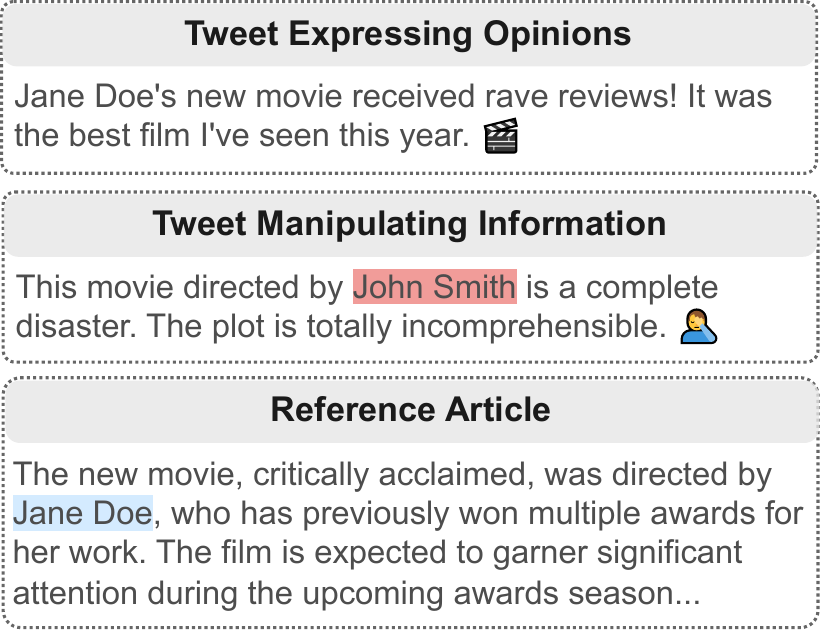}
    \vspace{-2mm}
    \caption{Two illustrative examples that highlight the challenge of identifying manipulation of news on social media. The first example expresses a personal opinion about watching a well-reviewed movie without distorting any facts from the associated article. Conversely, in the second example, the tweet falsely asserts that the movie is directed by \hlc{lightred}{John Smith} instead of \hlc{lightblue}{Jane Doe}, thereby misrepresenting the information contained in the reference article. Hence, the second tweet misrepresents the information contained in the reference article. \looseness=-1} %
    \label{fig:toy_example}
    \vspace{-5mm}
\end{figure}
\section{Introduction}

Detecting texts that contain misrepresentations of information originally presented in reference texts is crucial for combating misinformation.
Previous research has primarily tackled this issue in the context of fact-checking \cite{thorne-etal-2018-fever, wadden-etal-2020-fact}, where the goal is to debunk unsupported claims using relevant passages, in fact-checking \cite{kryscinski-etal-2020-evaluating, fabbri-etal-2022-qafacteval, qiu-etal-2024-amrfact}, and in chart captioning \cite{huang-etal-2024-lvlms} where the focus is on assessing the faithfulness of generated summaries to the reference articles. However, none of the previous work has specifically addressed the identification of social media posts that manipulate information which was presented with a reference article from a news corpus. 
This poses a significant challenge due to \textit{the prevalence of personal opinions in social media posts}. Our experiments demonstrate that state-of-the-art fact-checking and faithfulness assessment frameworks do not yield high performance in identifying social media posts that manipulate information (see \Cref{sec:results}). To effectively tackle this problem, models must be able to discern between personal opinions and sentences that distort information in social media posts. Examples of tweets that only express personal opinions and tweets that manipulate information can be found in \Cref{fig:toy_example}. %

In this paper, we introduce a new task called \textit{identifying manipulation of news on social media}. Given a social media post and its associated news article, models are tasked to understand whether and how the post manipulates information presented in the article. We define \textit{manipulation} as cases where \textit{a social media post intentionally misrepresents and distorts the content of the reference article}, following prior relevant studies \cite{shuetal2017fake, fung-etal-2021-infosurgeon}.  To explore this problem, we repurposed news articles from FakeNewsNet \cite{shu2020fakenewsnet} and constructed a fully-annotated dataset, \datashort~, consisting of 3.6K tweets accompanied by
their corresponding news articles. %
To improve annotation cost-efficiency, we propose a two-stage data collection pipeline instead of naively requesting annotators to annotate a subset of human-written tweets from \fakenewsnet~. This approach tackles imbalanced tweet distributions, where the majority of tweets do not manipulate the associated article. It also addresses the challenge of verifying information between news articles and tweets, making the annotation process more efficient. In the first round, human annotators are assigned the task of validating tweets generated by large language models (LLMs) in a controllable manner. The data collected from these rounds is subsequently utilized to train a sequence-to-sequence model for identifying manipulation within tweets authored by humans. In the second round of annotation, these human-authored tweets are labeled accordingly. The 0.5K human-written tweets annotated in the second round are used as the test set for evaluation. Conversely, the 3.1K machine-generated tweets collected in the first round are used for our training and development set.  \looseness=-1

Our study aims to address three main research questions. First, we investigate the comparison between the fine-tuning paradigm and the in-context learning paradigm for this task. Using our curated dataset, we evaluate the performance of the fine-tuned sequence-to-sequence model discussed earlier in comparison to state-of-the-art LLMs. Surprisingly, we discover that our \textbf{much smaller fine-tuned model outperforms LLMs prompted with zero-shot or few-shot exemplars on the proposed task}. In fact, we find that LLMs do not achieve satisfactory performance on our task when only provided with a few exemplars. Second, we explore the impact of various attributes of a news article on its susceptibility to manipulation. To conduct this analysis, we employ the previously described sequence-to-sequence model to analyze a vast collection of over 1M tweets and their associated articles. Our findings reveal \textbf{a higher likelihood of manipulation in social media posts when the associated news articles exhibit low trustworthiness or pertain to political topics}. Finally, we investigate the role of manipulated sentences within a news article. To address this question, we perform discourse analysis on the test set of \datashort~. Through this analysis, we uncover that \textbf{manipulated sentences within a news article often encompass the primary narrative or consequential aspects of the news article}.

Our contributions can be summarized as follows:
\begin{itemize}[noitemsep,nolistsep]
  \item We introduce and define the new task of identifying manipulation of news on social media. 
  \item We propose a novel annotation scheme for this task. Using this scheme, we construct a dataset consisting of 3.6K samples, carefully annotated by human experts. %
  \item We demonstrate that this dataset serves as a rigorous testbed for tackling identification of manipulation in social media. Specifically, we showcased the inadequate performance of LLMs in effectively addressing this challenge. \looseness=-1
  \item Our proposed framework combines an LLM with a smaller fine-tuned model, utilizing opinion sentences extracted by the LLM as additional features. This achieves the best performance for our task. \looseness=-1
\end{itemize}

\section{Identifying Manipulation of News on Social Media}

The goal of our task is to identify whether a social media post misrepresents information and what information is being manipulated given the associated reference article. Following prior work \cite{shuetal2017fake, fung-etal-2021-infosurgeon}, we define the term \textit{manipulation} as 
\newtheorem{definition}{Definition}
\begin{definition}
    A social media post is deemed to manipulate information when it intentionally misrepresents and distorts the content of the reference article.\looseness=-1
\end{definition}
The models are tasked to understand whether a tweet manipulates information in the reference article (\Cref{subsec:subtask1}), which newly introduced information in the tweet is used for manipulation (\Cref{subsec:subtask2}), and which original information in the reference article is manipulated (\Cref{subsec:subtask3}). In the following subsections, we provide detailed task formulation for each sub-task. \looseness=-1

\subsection{Sub-task 1: Tweet Manipulation Detection}
\label{subsec:subtask1}
Given a tweet and its associated news article, the first subtask is to classify the manipulation label $l$ of this tweet, where $l \in \{\recontextualized~, \notrecontextualized~ \}$. %
A tweet is considered \recontextualized~ as long as there is at least one sentence that comments on the content of the associated article, and this sentence contains manipulated or inserted information. Otherwise, this tweet is \notrecontextualized~.

\subsection{Sub-task 2: Manipulating Span Localization}
\label{subsec:subtask2}
Once a tweet is classified as \recontextualized~, the next step is determining which information in the reference article was manipulated in the tweet. We refer to the information being manipulated as the \textit{pristine span}, and the newly introduced information as the \textit{manipulating span}. Both \textit{pristine span} and \textit{manipulating span} are represented as a text span in the reference article and the tweet, respectively. Identifying both information can help provide interpretability on model outputs and enable finer-grained analysis that provides more insights, as demonstrated in \Cref{sec:exploratory_analysis}. Using \Cref{fig:toy_example} as an example, the \textit{manipulating span} is \hlc{lightred}{\textit{John Smith}}. \looseness=-1 %

\subsection{Sub-task 3: Pristine Span Localization}
\label{subsec:subtask3}

Similar to the second task, in this task, the model should output the \textit{pristine span} that is being manipulated. In cases where the \textit{manipulating span} is simply inserted, and no \textit{pristine span} is manipulated, models should output a null span or an empty string. Using \Cref{fig:toy_example} as an example, the \textit{pristine span} is \hlc{lightblue}{\textit{Jane Doe}}. 

\section{The \datashort~ Dataset}
\label{sec:data}
 
 Our dataset consists of 3,636 tweets associated with 2,688 news articles. Each sample is annotated with (1) whether the tweet manipulates information presented in the associated news article, (2) which new information is being introduced, and (3) which information is being manipulated. We refer to this dataset as the \datashort~ dataset. An overview of the data curation process is shown in \Cref{fig:data_curation_process}. The following sections describe our corpus collection and annotation process.

\subsection{News Article Source}

To facilitate the analysis of human-written tweets, we created \datashort~ by repurposing a fake news detection dataset, \fakenewsnet~ \cite{shu2020fakenewsnet}. \fakenewsnet~ contains news articles from two fact-checking websites, \politifact~ and \gossipcop~, where each news article is annotated with a factuality label. In addition, for each news article, \fakenewsnet~ also consists of user engagement data, such as tweets, retweets, and likes, on Twitter. We reused the news content and the associated tweets from \fakenewsnet~ for our \datashort~ dataset. %

During the early stage of the experiment, we observe that some news articles in \fakenewsnet~ are inappropriate for our study due to insufficient textual context. For example, some articles only contain a news title, a video, and a caption. To avoid such content, we remove news pieces containing less than 300 tokens.

\subsection{Tweet Collection}
Creating a high-quality dataset for our task using human annotators is extremely expensive and time-consuming primarily because the annotation task is challenging. Furthermore, real-world tweets authored by humans typically do not manipulate the associated articles. To address these issues, we have devised a two-stage pipeline to create training data.
In the first round of annotation, we utilize ChatGPT\footnote{\texttt{GPT-3.5-turbo}} to generate both \recontextualized~ and \notrecontextualized~ tweets in a controllable manner. Human annotators are then tasked with validating the generated tweets for their validity (\Cref{subsec:tweet_gen}). In the second round of annotation, we train a model on the data collected from the previous two rounds and employ this model to identify \recontextualized~ human-written tweets for human annotation (\Cref{subsec:data_human_val}). This approach ensures that annotators are not overwhelmed with a large number of \notrecontextualized~ tweets, resulting in significant improvements in time and cost efficiency compared to the aforementioned naive method. \looseness=-1

\subsubsection{Tweet Generation}
\label{subsec:tweet_gen}
We first used Stanza to extract \textsc{Location}, \textsc{People}, and \textsc{Event} named entities from all news articles. Then, we prompted ChatGPT to generate \notrecontextualized~ and \recontextualized~ tweets for each news article. The span of these entities are denoted as $S = \{S_0, S_1, ..., S_n\}$. The prompts used for generating these tweets are as follows:

\begin{quote}
\textbf{\notrecontextualized}: \texttt{This is a news article: \textbf{NEWS\_ARTICLE}. Write a tweet that comments on this article. Keep it within 280 characters:}
\end{quote}

\begin{quote}
\textbf{\recontextualized}: \texttt{This is a news article: \textbf{NEWS\_ARTICLE}. Write a tweet that comments on this article but changes \hlc{lightblue}{\textbf{PRISTINE\_SPAN}} to \hlc{lightred}{\textbf{NEW\_SPAN}} and includes NEW\_ENTITY in your tweet. Keep it within 280 characters: }
\end{quote}
Here, \hlc{lightblue}{\texttt{PRISTINE\_SPAN}} is a span randomly sampled from the spans of all named entities belonging to \texttt{NEWS\_ARTICLE} %
, whereas \hlc{lightred}{\texttt{NEW\_SPAN}} is another span sampled from $S$ with the same entity type as \hlc{lightblue}{\texttt{PRISTINE\_SPAN}}%
. We have also experimented with other prompt templates. While the overall generation quality does not differ much, these prompt templates most effectively prevent ChatGPT from generating undesirable sequences such as "\textit{As an AI language model, I cannot ...}". 

In addition to generating \recontextualized~ tweets where new information is manipulated from the original information contained in the associated article, we also produce \recontextualized~ tweets where new information is simply inserted into the tweet using the following prompt:

\begin{quote}
    \texttt{This is a news article: \textbf{NEWS\_ARTICLE}. Summarize the article into a tweet and comment about it. Include \hlc{lightred}{\textbf{NEW\_SPAN}} in your summarization but do not include \hlc{lightred}{\textbf{NEW\_SPAN}} in the hashtag\footnote{We instruct ChatGPT not to include \texttt{\hlc{lightred}{\textbf{NEW\_SPAN}}} in the hashtag. Otherwise, ChatGPT often does not insert \texttt{\hlc{lightred}{\textbf{NEW\_SPAN}}} into the main text of the tweet.}. Keep it within 280 characters:}
\end{quote}

To further improve data quality and reduce costs in human validation, we only keep \notrecontextualized~
tweets that contain at least one sentence inferrable from the corresponding article. Concretely, we use DocNLI \cite{yin-etal-2021-docnli}, a document-level entailment model, to determine the entailment probability between the reference article and each tweet sentence. A valid consistent tweet must have at least one sentence with an entailment probability greater than 50\%. Additionally, we remove \recontextualized~ tweets that do not contain the corresponding \hlc{lightred}{\texttt{NEW\_SPAN}} specified in the corresponding prompts.

While we initially considered using various prompts to generate tweets in order to achieve greater diversity, our early experiments revealed that the resulting outputs did not exhibit significant variations in terms of styles and formats. Furthermore, ChatGPT possesses the capability to produce tweets with diverse styles even when the same prompt template is used. As a result, we have chosen to use a single prompt for all experiments.

\subsubsection{Our Proposed Annotation Process}
\label{subsec:data_human_val}

We use Amazon’s Mechanical Turk (AMT) to conduct annotation. Annotators were provided with a reference article and a corresponding generated tweet, along with labels indicating whether the tweet manipulates the article, and whether the predicted \hlc{lightred}{\texttt{NEW\_SPAN}} and \hlc{lightblue}{\texttt{PRISTINE\_SPAN}} are accurate. In the first round of annotation, annotators were presented with tweets generated by ChatGPT. The labels for these tweets were naively derived from the data generation process, where we determined the manipulation label, \hlc{lightred}{\texttt{NEW\_SPAN}}, and \hlc{lightblue}{\texttt{PRISTINE\_SPAN}} before prompting ChatGPT to generate a tweet. For efficient annotation, the annotators only need to validate whether the labels derived from the ChatGPT prompts are correct. We keep samples whose labels for all three sub-tasks are correct, while the others are discarded.
In the second round of annotation, human-written tweets were annotated, and the predicted labels for these tweets were obtained from a model (see below paragraphs) trained on the data collected in the first annotation round. For detailed information regarding annotation guidelines and the user interface, please refer to \Cref{apx:annotation_details}. The following paragraphs provide an overview of our annotation process.

\begin{table}[t]
    \small
    \centering
    \begin{adjustbox}{max width=0.48\textwidth}
    {
    \begin{tabular}{ccccc}
        \toprule
        
        Split  & \# \recontextualized~ & \# \notrecontextualized~ & \# Doc & Tweet Author  \\
        \midrule
        Train  & 1,465 & 851 & 1,963 & Machine \\
        Dev  & 482 & 318 & 753 &  Machine \\
        Test  & 294 & 226 & 299 & Human \\
        \bottomrule
    \end{tabular}
    }
    \end{adjustbox}
    \vspace{-2mm}
    \caption{Statistics of our \datashort~ dataset. }
    \vspace{-5mm}
    \label{tab:dataset_stats}
    
\end{table}

\paragraph{First Round} 
The first round of annotation is for curating machine-generated tweets, which are used as our training set and development set. Initially, for annotator qualification, three annotators worked on each of our HITs\footnote{HIT refers to the Human Intelligence Task, which is the unit for an annotation task in Amazon Mechanical Turk.}. We used the first 100 HITs to train annotators by instructing them where their annotations were incorrect. Then, the next 100 HITs were used to compute the inter-annotator agreement (IAA). At this stage, we did not provide further instructions to the annotators. Using Fleiss' $\kappa$ \cite{fleiss1971measuring}, 
we obtain an average IAA of 62.4\% across all tasks, indicating a moderate level of agreement. Finally, we selected the top 15 performers as qualified annotators. These annotators were chosen based on how closely their annotations matched the majority vote for each HIT. \looseness=-1

Since the annotators have already been trained, we assigned each HIT to a single annotator to improve annotation efficiency for the remainder of the machine-generated tweets. In addition to being annotated by an MTurk worker, each annotation is also re-validated by a graduate student. The average agreement between the graduate student and the MTurk worker is 93.1\% per Cohen's $\kappa$ \cite{cohen1960coefficient}, implying a high agreement. We only keep samples where the validation done by the graduate student agrees with the annotation done by the worker. After two rounds of annotations, we collected 3,116 human-validated samples.\looseness=-1
\begin{figure}[t]
    \centering
    \includegraphics[width=0.9\linewidth]{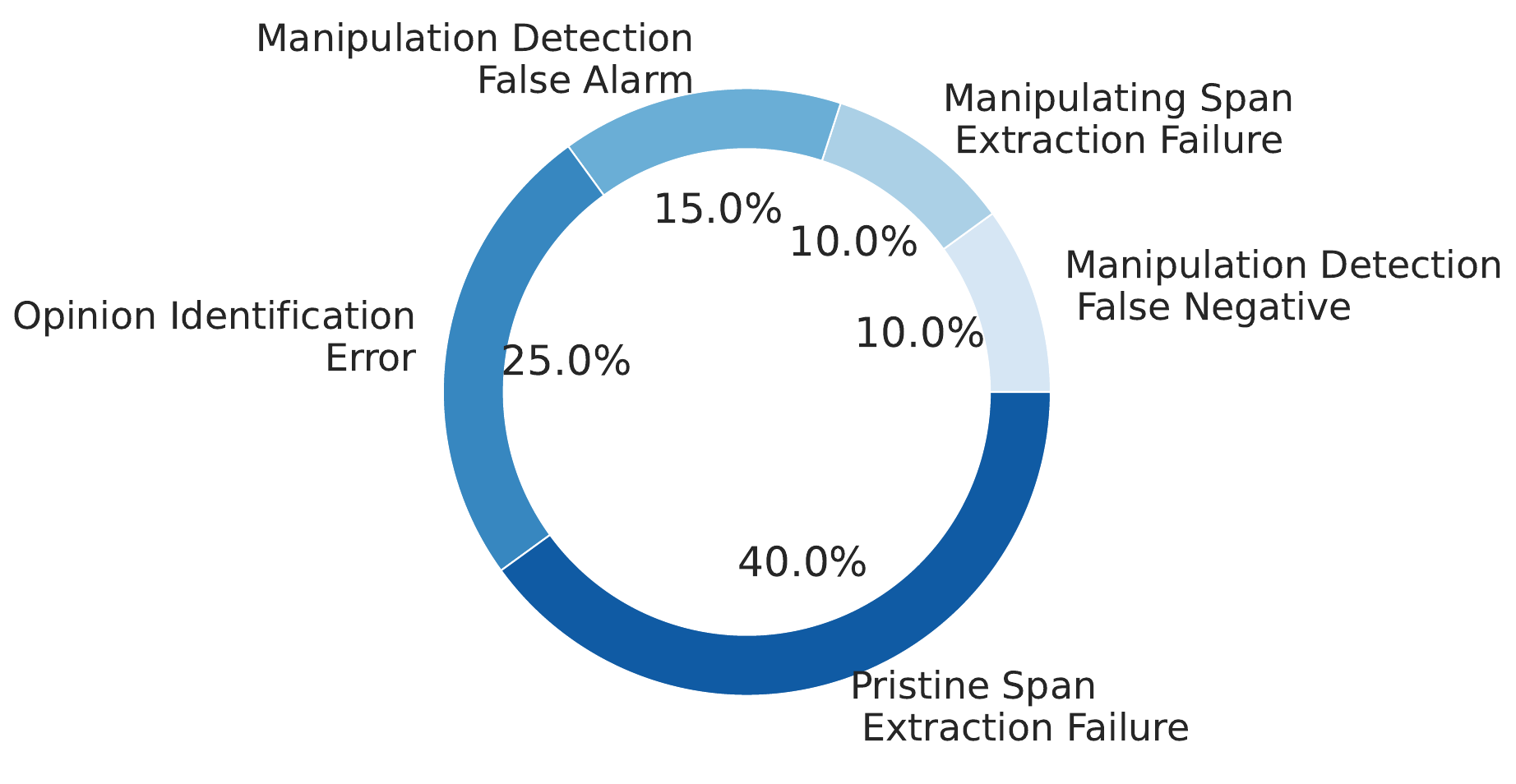}
    \vspace{-2mm}
    \caption{Distributions of errors. The error type definition is shown in \Cref{apx:error_type_defintion}.} %
    \vspace{-7mm}
    
    \label{fig:error_analysis}
\end{figure}

\paragraph{Second Round}
Using the 3K examples we collected, we train a sequence-to-sequence (seq2seq) model that learns to tackle all three tasks jointly. Concretely, we split the collected data into 2,316: 800 for training and validation. Model details are described in the next paragraph. Once the model was trained, we applied it to identify manipulation in the human-written tweets that are associated with the articles in FakeNewsNet. Then, we randomly sampled from predicted \recontextualized~ and  \notrecontextualized~ examples to be further validated by MTurk workers. The inter-annotator agreement between the graduate student and the MTurk worker is 73.0\% per Cohen's $\kappa$ \cite{cohen1960coefficient}. While the agreement is moderately high, it is much lower than that in the previous round. This suggests that manipulation in human-written tweets is more challenging to identify. The user interface of each round of annotation is shown in \Cref{apx:annotation_ui}. Finally, we have curated the \datashort~ dataset. The dataset statistics are shown in \Cref{tab:dataset_stats}.

\paragraph{Baseline Model}
In this paragraph, we describe the model we used to facilitate the second round of annotation. Motivated by the advantages of generative models over sequence-tagging models \cite{li-etal-2021-document, huang-etal-2021-document, hsu-etal-2022-degree}, we trained a seq2seq model based on LongFormer-Encoder-Decoder (LED)\footnote{\url{https://huggingface.co/allenai/led-base-16384}} \cite{beltagy2020longformer} that learns to solve the three tasks jointly. We name this model \textbf{LED-FT}.

Formally, the input $x = [t\|a]$ to our model is the concatenation of a tweet $t$ and the corresponding article $a$. The objective of the model is maximum likelihood estimation, 
\begin{align}
    \mathcal{L} = - \sum_i p(y_i | y_{<i}, x),
\end{align}
where $y_i$ denotes the $i$-th token in the decoding targets. Concretely, if the article is \notrecontextualized~, the model should output ``\texttt{No manipulation}''. Otherwise, the model should output
``\texttt{\textbf{Manipulating span}: \hlc{lightred}{NEW\_SPAN} \textbackslash ~ \textbf{Pristine span}:  \hlc{lightblue}{PRISTINE\_SPAN}}''. For cases where \texttt{\hlc{lightred}{NEW\_SPAN}} is merely inserted into the tweet, the model will output ``None'' for \texttt{\hlc{lightblue}{PRISTINE\_SPAN}}. Details of inputs, outputs, and training hyper-parameters can be found in \Cref{apx:training_details}.%

\begin{figure}[t]
    \centering
    \includegraphics[width=0.65\linewidth]{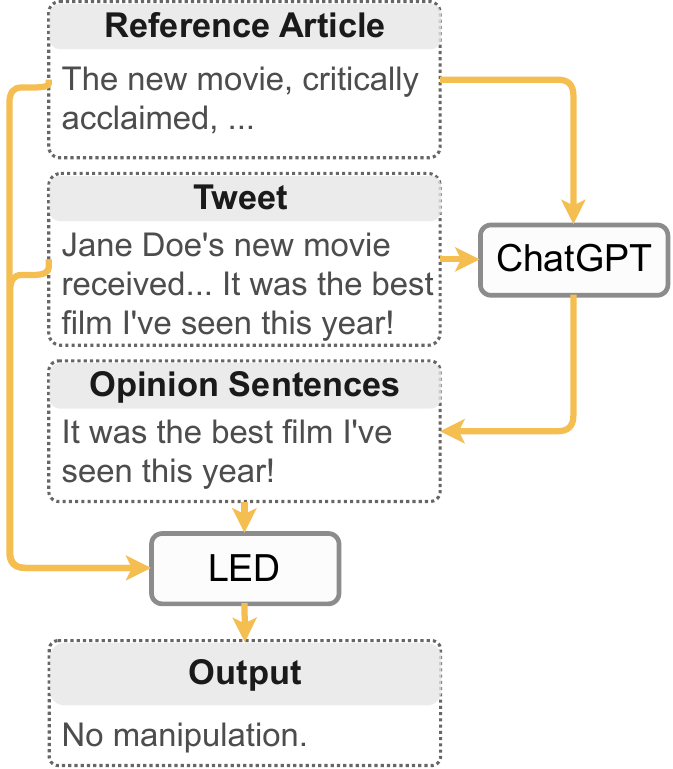}
    \vspace{-2mm}
    \caption{An overview of the proposed framework, \textbf{LLM + LED-FT}. We first use ChatGPT to identify sentences that express opinions from the tweet. Then, the opinion sentences are fed to a LED as additional features to help discern between sentences that express personal opinions and sentences that manipulates information. \looseness=-1} 
    \vspace{-6mm}
    
    \label{fig:method}
\end{figure}
\section{Methodology}
\label{sec:method}

We conducted an error analysis on the \textbf{LED-FT} model discussed in the previous section. Our analysis revealed that a significant portion of errors occurred due to the model's inability to distinguish between tweet sentences that express personal opinions and those that manipulate information from the associated article, as depicted in \Cref{fig:error_analysis} (refer to \Cref{apx:remaining_challenges} for further details). To address this issue, we propose a pipeline approach that involves utilizing ChatGPT to identify personal opinions within the tweet. This extracted opinions 
is then incorporated into our seq2seq model during both training and testing stages. An overview of the framework is shown in \Cref{fig:method}.

More specifically, we denote the identified opinion sentences in the tweet $t$ as $o = p_{\text{LLM}}(t, a, d)$, where $d$ represents the instruction provided to ChatGPT for opinion identification. The input to our fine-tuned model becomes $x' = [t\|a\|o]$, and the loss function remains as MLE:

{
\small
\begin{align}
    \mathcal{L}' = - \sum_i p(y_i | y_{<i}, x').
\end{align}
}
By incorporating this framework, we aim to enhance the model's ability to differentiate between personal opinions and instances where information is manipulated from the associated article. We name this pipeline \textbf{LLM + LED-FT}.%

\begin{table*}[t]
    \small
    \centering
    \begin{adjustbox}{max width=\textwidth}
    {
    \begin{tabular}{lcccccccc}
        \toprule
        
        \multirow{2}{*}{\textbf{Model}} & \multirow{2}{*}{\textbf{Learning Method}} & \textbf{Sub-task 1} & \multicolumn{3}{c}{\textbf{Sub-task 2}} & \multicolumn{3}{c}{\textbf{Sub-task 3}} \\
        \cmidrule(lr){3-3}\cmidrule(lr){4-6}\cmidrule(lr){7-9}
        & & F1 & EM & F1 & RL & EM & F1 & RL \\ %

        \midrule
        Human & - & 89.92 & 44.23 & 67.93 & 68.82 & 42.88 & 65.29 & 66.31 \\
        \midrule

        Vicuna & Zero-shot & 47.09 & 1.35 & 5.11 & 6.07 & 4.04 & 6.21 & 7.06\\ %
        ChatGPT & Zero-shot & 52.49 & 1.54 & 13.30 & 15.96  & 4.42 & 7.46 & 8.35 \\
        ChatGPT & Two-shot ICL & 65.28 &  0.96 & 7.62 & 8.87 & 12.50 & 13.91 & 14.18 \\
        ChatGPT & Four-shot ICL & 54.69 & 3.07 & 12.79 & 15.15 & 1.54 & 4.99 & 5.95\\ 
        ChatGPT & Two-shot CoT & 52.92 & 1.54 & 7.70 & 9.21 & 4.42 & 5.86 & 6.12 \\
        ChatGPT & Four-shot CoT & 53.88 & 0.96 & 7.93 & 9.66 & 3.46 & 5.24 & 5.70\\
        \midrule 
        \textsc{Concrete} & Zero-shot & 57.88 &- & - & - & - & - & - \\
        DocNLI  & Zero-shot & 62.26 & - & - & - & - & - & - \\
        QAFactEval & Zero-shot & 62.56 & - & - & - & - & - & - \\
        
        \midrule 
        LED-FT (Ours) & Fine-tuned & 72.62$^*$ & 26.73$^*$ & 29.25$^*$ & 29.68$^*$ & 13.65$^*$ & 14.46 & 14.53 \\ 
        LLM + LED-FT (Ours) & Zero-shot + Fine-tuned & \textbf{73.46}$^*$ & \textbf{28.85}$^*$ & \textbf{31.72}$^*$ & \textbf{32.32}$^*$ & \textbf{15.19}$^*$ & \textbf{16.21}$^*$ & \textbf{16.41}$^*$ \\

        \bottomrule
    \end{tabular}
    }
    \end{adjustbox}
    \vspace{-2mm}
    \caption{Performance (\%) of different models on the \datashort~ test set. EM denotes Exact Match, and RL denotes ROUGE-L. Statistical significance over best-performing LLMs computed with the paired bootstrap procedure \cite{berg-kirkpatrick-etal-2012-empirical} are indicated with $^*$ ($p< .01$).} 
    
    \label{tab:main}
    \vspace{-6mm}
\end{table*}

\section{Experimental Setup}

\subsection{Evaluation Metrics}
Subtask 1 involves a binary classification problem, and thus, the Macro F1 score serves as the evaluation metric. For subtasks 2 and 3, in addition to Exact Match, we use Macro Overlap F1 score \cite{rajpurkar-etal-2016-squad} and ROUGE-L \cite{lin-2004-rouge} as the metrics to more accurately assess model performance by allowing models to receive partial credit for correctly identifying some parts of the information, even if they fail to output the entire text span.\looseness=-1 %

\subsection{Baselines}
\label{subsec:baselines}
We compare our proposed framework with various recently released large language models (LLMs), including Vicuna\footnote{Vicuna-13b is evaluated in our experiment.} \cite{vicuna2023} and ChatGPT, which have demonstrated superior language understanding and reasoning capabilities. ChatGPT is an improved version of InstructGPT \cite{ouyang2022training} that was optimized for generating conversational responses. On the other hand, Vicuna is a LLaMA model \cite{touvron2023llama} fine-tuned on ShareGPT\footnote{\url{https://sharegpt.com/}} data, and has exhibited advantages compared to other open-source LLMs, such as LLaMA and Alpaca \cite{alpaca}. We tested the zero-shot, two-shot, and four-shot performance of ChatGPT in both in-context learning (ICL) and chain-of-thought (CoT) \cite{wei2022chain} settngs , where the in-context exemplars are randomly chosen from our training set. For Vicuna, we only evaluated its zero-shot ability as we found that it often outputs undesirable texts when exemplars are provided. The details of our prompts for these LLMs can be found in \Cref{apx:llm_prompts}. In addition, we also evaluate one fact-checking framework, \textsc{Concrete} \cite{huang-etal-2022-concrete}, and two faithfulness evaluation frameworks, QAFactEval \cite{fabbri-etal-2022-qafacteval} and DocNLI \cite{yin-etal-2021-docnli} on our subtask 1. Similar to previous studies, we establish the faithfulness thresholds for both frameworks by selecting the values that yield the highest performance on our development set.%

\section{Results}
\label{sec:results}

\subsection{Performance on \datashort~}
\Cref{tab:main} presents a summary of the main findings from our evaluation on the \datashort~ test set. We have made several interesting observations: First, all LLMs we tested performed poorly across the three proposed tasks. This indicates that simply prompting LLMs, whether with or without exemplars, is not sufficient to effectively address the problem of identifying manipulation of news on social media. We also found that providing more exemplars do not work well on our task as the performance drop when we increase the number of in-context exemplars from 2 to 4. This is likely caused by the long-context nature of our task. Indeed, the average number of tokens per article is 2609.6 in the test set. Secondly, despite its simplicity and smaller size compared to the LLMs, \textbf{LED-FT} outperforms all baseline models significantly in identifying social media manipulation across all three tasks. This outcome highlights the value and importance of our training data and suggests that a fine-tuned smaller model can outshine larger models when tackling challenging tasks. Finally, the proposed \textbf{LLM + LED-FT} outperforms all other models, including \textbf{LED-FT} significantly. This implies that LLMs can complement smaller fine-tuned models by  identifying opinions and that the ability to identify opinion sentences from social media posts is critical for our task. Examples of how the opinions extracted by ChatGPT help correct errors can be found in \Cref{apx:qualitative_examples}. \looseness=-1

In order to gauge the feasibility of the task, we enlisted the assistance of a graduate student to tackle our test set. While this may not necessarily represent the upper bound of performance, it provides a preliminary approximation of human performance. As depicted in \Cref{tab:main}, there remains a discernible gap between \textbf{LLM + LED-FT} and human performance. This highlights great opportunities in our task for future research.

\begin{figure}[t]
    \centering
    \begin{subfigure}{0.2\textwidth}

    \includegraphics[width=.95\linewidth]{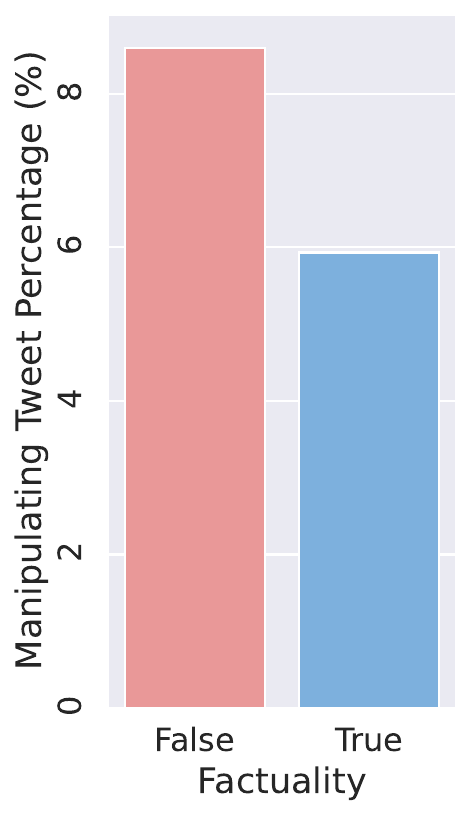}
    \vspace{-2mm}
    \end{subfigure}
    ~
    \begin{subfigure}[b]{0.2\textwidth}
    \includegraphics[width=.95\linewidth]{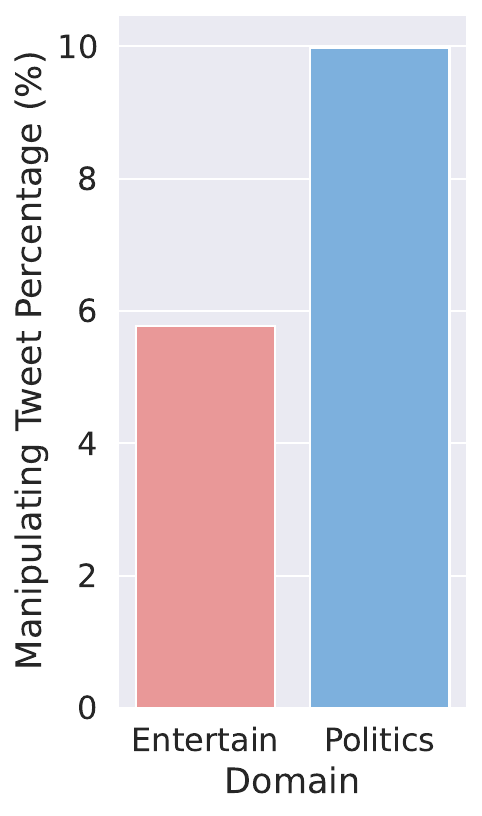}
    \vspace{-2mm}
    \end{subfigure}

    \caption{The percentage of tweets that manipulate the associated articles across different levels of factuality and domains.\looseness=-1}
    \label{fig:recontextualization_highlevel}
    \vspace{-5mm}
\end{figure}
\subsection{Exploratory Analysis}
The proposed \textbf{LED-FT} model enables us to perform a large-scale study of manipulation on the \datashort~ test set and the 1M human-authored tweets associated with the news articles from the FakeNewsNet dataset. In this section, we explore how an article is \recontextualized~ and how different properties of a news article, such as domain and factuality affect manipulation.

\label{sec:exploratory_analysis}

\paragraph{Insight 1: Low-trustworthiness and political news are more likely to be manipulated.}
\Cref{fig:recontextualization_highlevel} shows the percentage of the 1M human-written tweets that are manipulated across 
2 domains and factuality levels.\footnote{The domain and factuality labels of each news article are already annotated in the FakeNewsNet dataset.} We first observe that tweets associated with \textit{False} news are more likely to be manipulated. One possible explanation is that audience of low-trustworthy news media may pay less attention to facts. Hence, they are more likely to manipulate information from the reference article accidentally when posting tweets. In addition, we also see that tweets associated with \textit{Politics} news are more frequently manipulated than those with \textit{Entertainment} articles. This could be explained by the fact that people have a stronger incentive to manipulate information for political tweets due to elections or campaigns.

\begin{figure}[bt]
    \centering
    \includegraphics[width=0.9\linewidth]{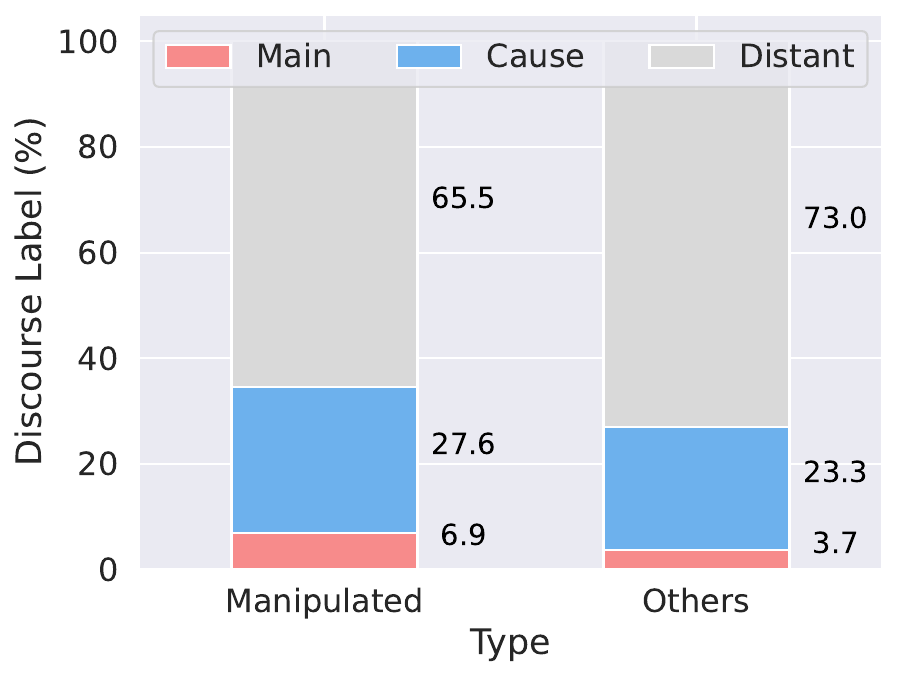}
    \vspace{-2mm}
    \caption{Results of discourse analysis. Manipulated sentences within news articles tend to encompass the main story (\textit{Main}) or convey the consequential aspects (\textit{Cause}) of the corresponding news story.\looseness=-1}
    \label{fig:discourse_results}
    \vspace{-6mm}
\end{figure}
\paragraph{Insight 2: Manipulated sentences are more likely to contain the main story or consequence of a news story.}
To discover the role of the sentence being manipulated in the reference article, we conducted discourse analysis on these sentences. We only conducted the analysis on our test set instead of the entire 1M human-written tweets for this analysis. Concretely, we formulate the discourse classification task as a sequence-to-sequence problem and train a LED-based model on the \textsc{NewsDiscourse} dataset \cite{choubey-etal-2020-discourse} using a similar strategy discussed in \Cref{subsec:data_human_val}. The learned discourse classification model achieves a Micro F1 score of 67.7\%, which is on par with the state-of-the-art method \cite{spangher-etal-2021-multitask}. Upon the discourse classification model being trained, we applied it to all the sentences in the reference article to analyze the discourse distribution. As shown in \Cref{fig:discourse_results}, compared to other sentences, sentences that were manipulated are much more likely to contain \textit{Main} or \textit{Cause} discourse, which corresponds to \textit{the primary topic being discussed} and \textit{the underlying factor that led to a particular situation}, respectively. Examples of the manipulated sentences with a \textit{Main} or \textit{Cause} discourse can be found in \Cref{apx:discourse_examples}. \looseness=-1

\section{Related Work}

\subsection{Faithfulness}

Faithfulness is often referred to as the factual consistency between the inputs and outputs. This topic has mainly been studied in the field of summarization. Prior work on faithfulness can be divided into two categories: evaluation and enhancement, the former of which is more relevant to our study. One line of faithfulness evaluation work developed entailment-based metrics by training document-sentence entailment models on synthetic data \cite{kryscinski-etal-2020-evaluating, yin-etal-2021-docnli, qiu-etal-2024-amrfact,DBLP:conf/acl/Chan0J23} or using traditional natural language inference (NLI) models at the sentence level \cite{laban-etal-2022-summac}. Another line of studies evaluates faithfulness by comparing information units extracted from the summaries and input sources using QA \cite{wang-etal-2020-asking, deutsch-etal-2021-towards, fabbri-etal-2022-qafacteval}. \looseness=-1

Our task differs from faithfulness evaluation in two key ways. Firstly, for our task to be completed effectively, models must possess the additional capability of distinguishing tweet sentences that relate to the reference article from those that simply express opinions. %
In contrast, models evaluating faithfulness only need to identify whether each sentence in the output is inferable from the input. Secondly, we require models to not only identify which original information is being manipulated by the new information, but also to provide interpretability as to why a tweet has been manipulated. %

\subsection{Fact-checking}
Fact-checking is a task that determines the veracity of an input claim based on some evidence passages. Some work assumes the evidence candidates are provided, such as in the \textsc{FEVER} dataset \cite{thorne-etal-2018-fever} and the \textsc{SciFact} dataset \cite{wadden-etal-2020-fact}. Approaches for this category of fact-checking tasks often involve a retrieval module to retrieve relevant evidence from the given candidate pool, followed by a reasoning component that determines the compatibility between a piece of evidence and the input claim \cite{yin-roth-2018-twowingos, pradeep-etal-2021-scientific}. Other work focuses on the \textit{open-retrieval} setting, where evidence candidates are not provided, such as in the \textsc{Liar} dataset \cite{wang-2017-liar} and the \textsc{X-Fact} dataset \cite{gupta-srikumar-2021-x}. Recent methods~\cite{DBLP:journals/corr/abs-2402-07401,DBLP:journals/corr/abs-2410-20140} introduce multi-agent debate frameworks to enhance the fact-checking capability of LLMs. 
For this task formulation, one of the main challenges is to determine where and how to retrieve evidence. Some approaches determine the veracity of a claim based solely on the claim itself and the information learned by language models during the pre-training stage \cite{lee-etal-2021-towards}, other methods leverage a retrieval module to look for evidence on the internet \cite{gupta-srikumar-2021-x} or a set of trustworthy sources \cite{huang-etal-2022-concrete}. Similar to the faithfulness task, the key distinction between fact-checking and our proposed task lies in the additional requirement for models to possess the capability of discerning between tweet sentences that pertain to the reference article and those that merely express opinions.

\section{Conclusion}
In this study, we have introduced and defined a novel task called \textit{identifying manipulation of news on social media}, which aims to determine whether and how a social media post manipulates the associated news article. To address this challenge, we meticulously collected a dataset named \datashort~, composed of both human-written and machine-generated tweets. Our analysis revealed that existing large language models (LLMs) prompted with zero-shot and two-shot exemplars do not yield satisfactory performance on our dataset, highlighting avenues for future research. We believe that the resources presented in this paper can serve as valuable assets in combating the dissemination of false information on social media, particularly in tackling the issue of news manipulation. \looseness=-1

\section{Limitations}

\paragraph{Using LLMs for data creation.}
 LLMs, such as ChatGPT, are instrumental in crafting entire tweets that are not only coherent but also conditioned on the specifics of the given news article, ensuring a level of fluency that mimics that of human writers. Moreover, the tweets fashioned by ChatGPT showcase a distinct superiority in quality when compared to more traditional methods of data synthesis, such as those that are rule-based or template-based. These earlier approaches often resulted in output that was both stilted and monotonous, falling short in fluency and variety, a fact substantiated by references\cite{goyal-durrett-2021-annotating, utama-etal-2022-falsesum}. By leveraging the capabilities of ChatGPT, we can generate machine-authored tweets that not only boast a broad diversity but also maintain a convincingly realistic quality, thereby providing an enriched dataset for scalable human annotation.

\paragraph{LLM prompts.} In our experiments involving prompting LLMs, we only explored ICL and CoT for prompting LLMs. There is a possibility that LLMs can achieve better performance when provided with more in-context exemplars and when prompted in a more refined manner. \looseness=-1

\section{Ethical Considerations}
The primary ethical consideration in our work pertains to the presence of false information in two aspects: tweets that manipulate the associated news articles and the inclusion of false news from the FakeNewsNet dataset. As with other fact-checking and fake news detection research, it is important to acknowledge the dual-use concerns associated with the resources presented in this work. While our resources can contribute to combating false information, they also possess the potential for misuse. For instance, there is a risk that malicious users could utilize the manipulating tweets or fake news articles to train a text generator for creating deceptive content. We highlight appropriate and inappropriate uses of our dataset in various scenarios: \looseness=-1

\begin{itemize}
    \item \textbf{Appropriate}: Researchers can use our framework to study the manipulation issue on social media and develop stronger models for identifying social media posts that manipulate information.
    \item \textbf{Inappropriate}: The fake news and manipulating tweets in \datashort~ cannot be used to train text generators for malicious purposes.
    \item \textbf{Inappropriate}: Use the  manipulation prompts discussed in this paper to generate tweets and spread false information.
    \item \textbf{Inappropriate}: The fake news in \datashort~ should not be used as evidence for fact-checking claims.
\end{itemize}

Furthermore, the privacy of tweet users is another aspect that warrants consideration, given that we are releasing human-written tweets. However, we assure that the dataset does not pose significant privacy concerns. The tweets in our dataset are anonymized, and it is important to note that all the associated news articles were already publicly available. Therefore, the release of this dataset should not have adverse implications for privacy.

\section*{Acknowledgement}
This research was done with funding from the Defense Advanced Research Projects Agency (DARPA) under Contracts No. HR001120C0123 and HR0011-24-3-0325. The views, opinions, and/or findings expressed are those of the authors and should not be interpreted as representing the official views or policies of the Department of Defense or the U.S. Government.

\bibliography{anthology,custom}
\bibliographystyle{acl_natbib}
\clearpage
\appendix

\section{Additional Discussios}
\paragraph{If real-world tweets typically do not manipulate associated articles ( \Cref{sec:data}), how practical and relevant is the proposed task?}
While manipulated tweets that distort information from news articles may not be extremely common on social media, they can still have an outsized impact when they do occur. Even a small number of tweets that deliberately misrepresent the facts around a news story have the potential to spread wildly on social media and shape public discourse \cite{allcott2017social, starbird2017examining}. We would argue that the harm caused by manipulated tweets warrants research efforts into detecting and combating them, even if the absolute number of such tweets is low. A few viral manipulated tweets can still reach millions of users and significantly skewed perceptions around news events and issues. Identifying and fact-checking these tweets is key to limiting the spread of misinformation.

\paragraph{Discrepancies between the training set and the test set.}
Despite our best efforts to minimize the gap between the training set and test set of \datashort~, some discrepancies remain due to the training set being generated by machines and the test set being produced by humans. This limitation is primarily attributed to budget constraints. In fact, synthetically generating training data is a common strategy in relevant fields where extensive human annotation poses significant challenges, such as fake news detection \cite{huang-etal-2023-faking, fung-etal-2021-infosurgeon} and factual inconsistency detection \cite{kryscinski-etal-2020-evaluating, utama-etal-2022-falsesum}. In the future, with additional resources, we aim to create an additional training set consisting entirely of human-written tweets. By comparing the performance of models trained on this human-written training set with those trained on the machine-generated training set, we can gain further insights. However, we wanted to emphasize that our test set exclusively consists of tweets authored by humans, which ensures the relevance of our techniques and dataset for real-world applications in handling tweets produced by actual Twitter users. While our data collection method may introduce discrepancies in the distribution between the training and test sets, the fundamental purpose of our dataset remains consistent: to investigate the manipulation of news articles on social media.

\paragraph{Manipulation types.} Our approach focuses on manipulations of three types of entities: \textsc{Location}, \textsc{People}, and \textsc{Event}. This approach may fail in cases where the manipulation is complex, beyond entity-level perturbations or involving multiple entities. However, it is important to highlight that following a meticulous examination of 100 manipulated examples from our dataset, we found that an overwhelming \textbf{85\% of them involve named entity manipulations only}. Through this analysis, we categorized manipulations based on their intent and the nature of the information distortion, identifying three additional manipulation types in addition to entity-level manipulation:

\begin{itemize}
    \item Misattribution of Quotes or Actions (10\%): Where social media posts attribute incorrect quotes or actions to individuals or entities not associated with them in the referenced news articles.
    \item Exaggeration/Understatement (3\%): Manipulations that inflate or diminish the severity or importance of the facts presented in the articles.
    \item Temporal Distortion (2\%): Tweets misleadingly suggest that certain events happened at a different time than reported in the article, affecting the perceived relevance or cause-effect relationships.
\end{itemize}

Based on this analysis, we have established stronger support for our claim in the paper and enriched our understanding of various manipulation types for future research. This highlights that our formulation is still relevant and can handle the vast majority of real-world manipulations.

\paragraph{How \texttt{\hlc{lightblue}{PRISTINE\_SPAN}} is mapped to \texttt{\hlc{lightred}{NEW\_SPAN}}?} \texttt{\hlc{lightblue}{PRISTINE\_SPAN}} refers to a text span within the reference article that is associated with a particular named entity and is relevant to the news narrative. \texttt{\hlc{lightred}{NEW\_SPAN}}, on the other hand, is a different text span associated with the same type of entity but is randomly sampled from the set of all named entities extracted from the news articles.

The intention behind replacing \texttt{\hlc{lightblue}{PRISTINE\_SPAN}} with \texttt{\hlc{lightred}{NEW\_SPAN}} is to create a manipulated piece of text by altering entity-related information found in the original article. By ensuring that the \texttt{\hlc{lightred}{NEW\_SPAN}} shares the same entity type as the \texttt{\hlc{lightblue}{PRISTINE\_SPAN}}, we maintain the semantic plausibility of the generated tweet.

For example, consider the following:

\begin{quote}
    \textbf{Reference Article}: ``President Smith advocated for environmental policies in the recent summit held in Geneva, emphasizing the need for sustainable development.'' (\texttt{\hlc{lightblue}{PRISTINE\_SPAN}}: ``President Smith'')        
\end{quote}

By extracting named entities, we might get a list like [``President Smith'', ``Geneva'', ``Prime Minister Johnson'', ``Paris'']. Suppose we choose ``Prime Minister Johnson'' as the \texttt{\hlc{lightred}{NEW\_SPAN}} to replace ``President Smith''. The manipulating tweet could then be:

\begin{quote}
    \textbf{Manipulating Tweet}: ``Prime Minister Johnson pushed for new economic measures in the conference that took place in Paris, expressing urgency for financial reform.'' (\texttt{\hlc{lightred}{NEW\_SPAN}}: ``Prime Minister Johnson'')    
\end{quote}

Here, the \texttt{\hlc{lightred}{NEW\_SPAN}} provides alternative, yet topically coherent, entities to create misinformation while preserving the sentence structure and general subject matter of the original article.

\paragraph{The prompts given to ChatGPT are pretty lengthy and may not be well articulated to the desired answers, and more shots given even result in worse performance.}

Our study aimed to explore the baseline effectiveness of LLMs such as ChatGPT and Vicuna in the task of identifying news manipulation on social media without extensive prompt engineering. This choice was deliberate to mirror a more \textit{generalizable and accessible} use case, where users of varying technical backgrounds rely on LLMs.

The prompts were carefully designed to reflect the task's complexity, ensuring clarity in instructions to produce relevant and accurate responses. Our aim was not to maximize the performance through prompt engineering but to \textbf{establish a fundamental understanding of LLM capabilities in this novel task domain under relatively straightforward conditions}.

To clarify, the drop in performance with more in-context examples suggests that this task likely requires additional abilities beyond simply providing more examples, which is an insightful result in itself, indicating areas for future research in improving LLMs' handling of complex and long-context relations in texts.

\paragraph{Is it true that the unsatisfying performance of LLMs is due to the capability of the language model or the prompt engineering?}
We tested models that have stronger long-context reasoning ability, such as GPT-4 (with a context window of 8K tokens). If these models show increased performance compared to ChatGPT and Vicuna, we can better conclude that the poor performance of ChatGPT and Vicuna is caused by their insufficient long-context reasoning abilities. In \Cref{tab:main_gpt4}, we show the performance of GPT-4 and GPT-4 Turbo on our task. Based on our findings, we can confirm that models with stronger long context reasoning ability are better at identifying manipulating tweets as well as  manipulated and inserted information. This validates our hypothesis that the poor performance of ChatGPT and Vicuna is caused by the long-context nature of our task and their limited ability in modeling long-form texts.

\paragraph{Are some entities more difficult to identify than others?}

\begin{table}[t]
    \small
    \centering
    \begin{adjustbox}{max width=0.5\textwidth}
    {
     \begin{tabular}{lccc}
        \toprule
        \textbf{Model} & \textbf{Person (\%)} & \textbf{Location (\%)} & \textbf{Event (\%)} \\
        \midrule 
        ChatGPT Two-shot ICL & 64.5 & 58.68 & 68.14 \\
        LED-FT (Ours) & 71.01 & 66.46 & 73.16 \\
        LLM + LED-FT (Ours) & 73.21 & 72.21 & 72.33 \\
        \bottomrule
    \end{tabular}
    }
    \end{adjustbox}
    \vspace{-2mm}
    \caption{Breakdown F1 scores w.r.t. different entity types.} 
    
    \label{tab:score_breakdown}
    
\end{table}

We ran an additional analysis to understand the performance breakdown for each error type. The results are summarized in the \Cref{tab:score_breakdown}.

Overall, we can see that manipulation of location-related entities is the most challenging to identify. We also found that by utilizing opinion sentences identified by LLM, we achieve significant performance gain on manipulations involving Person and Location entities. This highlights the effectiveness of the proposed framework.

\begin{table}[t]
    \small
    \centering
    \begin{adjustbox}{max width=0.5\textwidth}
    {
    \begin{tabular}{lcccc}
        \toprule
        
        \textbf{Model} & \textbf{Prompts} & \textbf{Sub-task 1} & \textbf{Sub-task 2} & \textbf{Sub-task 3} \\

        \midrule
        GPT-4 & Zero-shot  & 70.23 & 22.92 & 10.56 \\
        GPT-4 Turbo & Zero-shot & 72.21 & 19.56 & 12.43 \\ 
        
        \bottomrule
    \end{tabular}
    }
    \end{adjustbox}
    \vspace{-2mm}
    \caption{F1 scores of GPT-4 and GPT-4 Turbo on the \datashort~ test set.} 
    
    \label{tab:main_gpt4}
    
\end{table}

\section{Training Details}
\label{apx:training_details}
\subsection{LED-based Fine-tuned Model}
The input to our LED-based model is a concatenation of a tweet and a reference article: 
\begin{quote}
    \texttt{\textbf{Tweet}: TWEET \textbackslash \\
    \textbf{Reference article}: REF\_ARTICLE}
\end{quote}
If the article is \notrecontextualized~, the model should output:
\begin{quote}
    \texttt{No manipulation}
\end{quote}
Otherwise, the model should output the following:
\begin{quote}
    \texttt{\textbf{Manipulating span}: \hlc{lightred}{NEW\_SPAN} \textbackslash \\
    \textbf{Pristine span}: \\ \hlc{lightblue}{PRISTINE\_SPAN}}
\end{quote}

For cases where \texttt{\hlc{lightred}{NEW\_SPAN}} is merely inserted into the tweet, the model will output ``None'' for \texttt{\hlc{lightblue}{PRISTINE\_SPAN}}. Using this formulation, our model is learned to optimize the maximum likelihood estimation loss. We set identical weights for all tokens in the outputs.   %
\subsection{ChatGPT Prompts}
The prompt to ChatGPT for identifying opinions is as follows:
\begin{quote}
    \texttt{\textbf{Tweet}: TWEET \textbackslash \\
    \textbf{Reference article}: REF\_ARTICLE}\\
    Given the above tweet and article. List the sentences in the tweet that merely express opinions instead of manipulating information from the article. If there is none, answer "None". Do not provide explanations.
\end{quote}
\subsection{Training Hyper-parameters}
To learn the model, we use a learning rate of 5e-5. The maximum input and output sequence length are 1024 and 32 tokens, respectively. The model is optimized using the AdamW optimizer \cite{loshchilov2018decoupled} with a batch size of 4 and a gradient accumulation of 8. During inference time, we use beam search as the decoding method with a beam width of 4.

\subsection{Training Discourse Analysis Model}
For this discourse analysis model, the input is a concatenation of the reference article and a sentence from the same reference article, while the output is one of the discourse labels defined in \textsc{NewsDiscourse}. We then compare the discourse label distribution for sentences that contain text span (\texttt{\hlc{lightblue}{PRISTINE\_SPAN}}) that are manipulated by a tweet versus that for other sentences, as shown in \Cref{fig:discourse_results}. 

\begin{figure*}[t]
    \centering
    \includegraphics[width=0.9\textwidth]{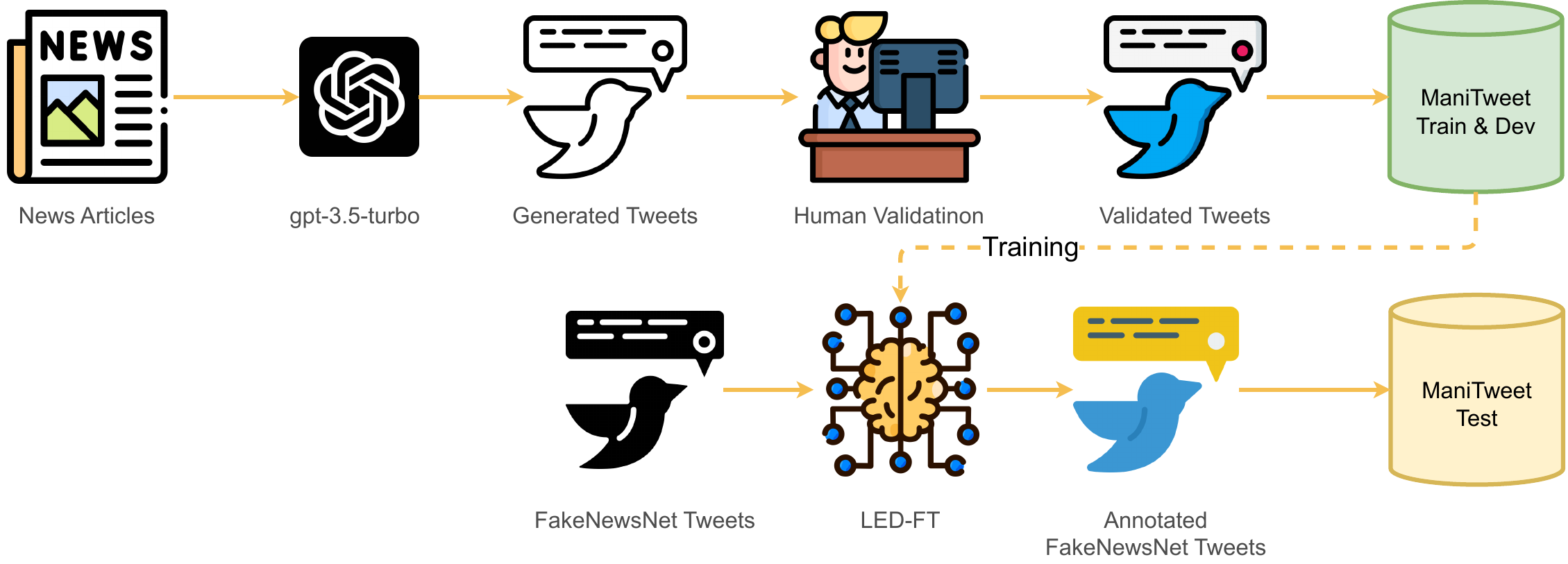}
    \vspace{-2mm}
    \caption{An overview of our data curation process.\looseness=-1} 
    \label{fig:data_curation_process}
    \vspace{-5mm}
\end{figure*}
\section{Error Analysis}
\label{apx:remaining_challenges}
To gain insights into the additional modeling and reasoning capabilities required for effectively addressing the task of social media manipulation, we manually compare 50 errors made by the LED-based model with ground-truth labels and analyze the sources of errors. The distribution of errors is illustrated in Figure \ref{fig:error_analysis}. Notably, the most prevalent error arises from the model's inability to extract the correct pristine span from the reference article that underwent manipulation. Among the 18 erroneous predictions in this category, 16 cases result from the model producing an empty string. This indicates that the model considers the manipulating information to be inserted when, in reality, it is manipulated from the information present in the reference articles. This could be attributed to the presence of 368 instances where the original information is an empty string, while the alternative answers for the original information only occur 1-2 times in other instances. This can be solved by scaling down the loss for these samples with an empty string as the label for original information. Additionally, another common type of error involves the model's failure to identify opinions expressed in the tweet. In these instances, the model considers the tweet to be manipulating information from the article, whereas the tweet primarily expresses opinions. Examples of these errors are presented in \Cref{apx:qualitative_examples}.
\section{Annotation Details}
\label{apx:annotation_details}
In this section, we describe the details of our annotation process. We show an overview of our data curation process in \Cref{fig:data_curation_process}. For better control of the annotation quality, we required that all annotators be from the U.S. and have completed at least 10,000 HITs with 99\% acceptance on previous HITs. The reward for each HIT
is \$1 U.S. dollar, complying with the ethical research standards outlined by AMT \cite{Salehi2015We}. Annotation interfaces are shown below.

\subsection{User Interface}
\label{apx:annotation_ui}
\begin{figure*}[b]
    \centering
    \includegraphics[width=0.9\linewidth]{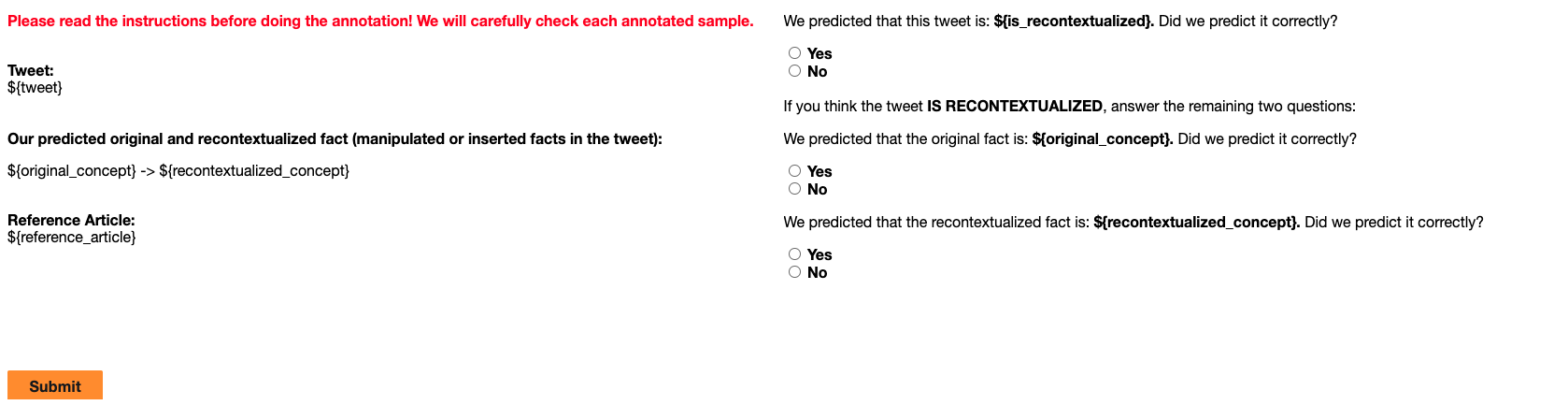}
    \vspace{-2mm}
    \caption{MTurk user interface for the first round of data annotation.}
    \vspace{-5mm}
    \label{fig:mturk_ui_2}
\end{figure*}

\begin{figure*}[b]
    \centering
    \includegraphics[width=0.9\linewidth]{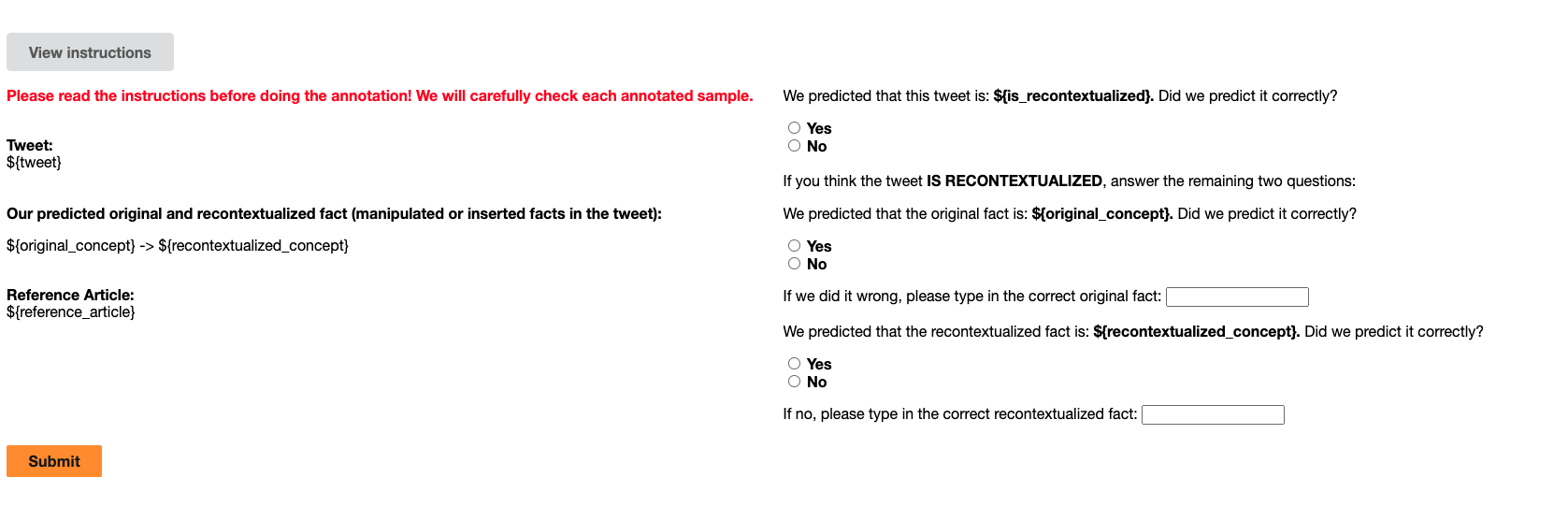}
    \vspace{-2mm}
    \caption{MTurk user interface for the second round of data annotation.}
    \vspace{-5mm}
    \label{fig:mturk_ui_3}
\end{figure*}

\Cref{fig:mturk_ui_2} and \Cref{fig:mturk_ui_3} display the annotation interface for the first round and the third round of annotation, respectively. The only difference is that for the second round of annotation, we asked annotators to correct errors made by our basic model discussed in \Cref{subsec:data_human_val}. Samples that do not receive ``yes'' on all three questions for the first round of annotation will be discarded. The rationale behind this design stems from three key reasons: Firstly, the data for the first round of annotation is automatically generated, enabling a relatively cost-effective approach to discard invalid samples and generate new ones, as opposed to requesting annotators to correct errors. Secondly, the data generated in these two rounds is predominantly valid, which eliminates the need for annotators to rectify errors and consequently accelerates the annotation process. Lastly, in the second round of annotation, by instructing annotators to identify errors made by our model, we can effectively identify the challenges faced by the model.

\section{Prompts for LLMs}
The zero-shot and two-shot prompt template to LLMs for the experiments discussed in \Cref{subsec:baselines} is shown in \Cref{tab:llm_prompt}. The in-context exemplars for the two-shot experiments are randomly sampled from the training set of \datashort~.
\begin{table*}[t]
    \small
    \centering
    {
    \begin{tabular}{p{0.15\linewidth} p{0.8\linewidth}}
        \toprule
        \textbf{Role} & \textbf{Utterance} \\
        \midrule
        
        User & You are tackling a social manipulation problem. You will be giving a tweet and an article, and your task is to identify which information from the article is misrepresented by which information in the tweet. You should answer in the following format ``Manipulating span: {manipulating\_span} Pristine span: {pristine\_span}'' in a single line. Here, \{manipulating\_span\} is the new information introduced in the tweet and {original\_concept} is the original information in the article. If the tweet simply inserts information, \{original\_concept\} should be "None". If the tweet does not manipulate the article, answer ``No manipulation''. You do not need to output other information such as an explanation. You don't need to provide code. In the following utterances, you will be presented a pair of tweet and news article. \\
        \midrule
        LLM &  I understand the task. Please provide me with the tweet and the corresponding article, and I will identify the manipulation. \\
        \midrule
        \textcolor{lightgray}{\multirow{4}{*}{User}} &  \textcolor{lightgray}{Tweet: Sad news for Ben Affleck and Lindsay Shookus as they split after a year of dating. It looks like Ben has been getting cozy with Playboy model Shauna Sexton, but sources say it's just light and fun. Best of luck to all involved. \#BenAffleck \#LindsayShookus \#ShaunaSexton} \\
        & \textcolor{lightgray}{Article: Ben Affleck and Lindsay Shookus Split After a Year of Dating} 
        \textcolor{lightgray}{Ben Affleck and Lindsay Shookus have broken up after over a year of dating, a source tells ET...}\\
        & \textcolor{lightgray}{=======================================================================}\\
        & \textcolor{lightgray}{For this example, you should answer ``No manipulation''} \\
        \midrule
        \textcolor{lightgray}{LLM} &  \textcolor{lightgray}{No manipulation} \\
        \midrule
        \textcolor{lightgray}{\multirow{4}{*}{User}} &  \textcolor{lightgray}{Tweet: Meghan's Lush Malibu Home Worth \$16M, which comes with 8 bedrooms, 8 bathrooms, a private beach, a pool, a home cinema - they even want to spend an     extra \$1 million to decorate it to their own taste. Unfortunately Caboolture scores 0 here!}\\
        & \textcolor{lightgray}{Article: Take A Look Inside Prince Harry And Meghan Markle's Lush Malibu Home If you're set to marry royalty, like Meghan Markle you would expect to move into a palace, and live out your Disney Princess fantasy...}\\
        & \textcolor{lightgray}{========================================================================}\\
        & \textcolor{lightgray}{For this example, you should answer ``Manipulating span: Caboolture Pristine span: None''} \\
    
        \midrule
        \textcolor{lightgray}{LLM} &  \textcolor{lightgray}{Manipulating span: Caboolture Pristine span: None} \\
        \midrule
        \multirow{2}{*}{User} &  Tweet: \textit{Input Tweet} \\
                              &   Article: \textit{Input Article}   \\
        \bottomrule
    \end{tabular}
    }
    
    \caption{The zero-shot and two-shot prompt template to the LLMs for producing the results in \Cref{tab:main}. Texts marked in \textcolor{lightgray}{gray} indicate that they are only used in the two-shot experiments.}
    \label{tab:llm_prompt}
    
\end{table*}

\begin{table*}[t]
    \small
    \centering
    {
    \begin{tabular}{p{0.15\linewidth} p{0.8\linewidth}}
        \toprule
        \textbf{Role} & \textbf{Utterance} \\
        \midrule
        
        User & You are tackling a social manipulation problem. You will be giving a tweet and an article, and your task is to identify which information from the article is misrepresented by which information in the tweet. You should answer in the following format: ``Opinion sentences: opinion\_sentences  Rationale: rationale Manipulating span: manipulating\_span   Pristine span: pristine\_span''. Here, \{opinion\_sentences\} is the sentences in the tweet that express opinions, \{rationale\} is your reasoning process that determines whether the tweet manipulates the article, \{manipulating\_span\} is the new information introduced in the tweet and \{original\_concept\} is the original information in the article. If the tweet simply inserts information, \{original\_concept\} should be ``None''. If the tweet does not manipulate the article, answer “No manipulation”. You do not need to output other information such as an explanation. The tweet does not manipulate the article if it is simply expressing opinions. In the following utterances, you will be presented a pair of tweets and news articles.
         \\
        \midrule
        LLM &  I understand the task. Please provide me with the tweet and the corresponding article, and I will identify the manipulation. \\
        \midrule
        \multirow{4}{*}{User} &  Tweet: Ariana Grande’s New Lavender Hair Color Just Inspired Our Next Trip To The Salon ...\\
        &  Article: Ariana Grande Dyed Her Hair, And This Is Our Favorite Color Transformation Yet Ariana \\
        & Grande is giving us whiplash with her hairstyles lately, and we honestly love it. On July 18th, Grande took to Instagram to debut her latest hair transformation. She’s now sporting pastel lavender locks and good god (is a woman), it looks amazing on her... \\
        & ======================================================================= \\
        & For this example, you should answer “Opinion sentences: Ariana Grande’s New Lavender Hair Color Just Inspired Our Next Trip To The Salon”: Rationale: The tweet is only expressing opinion and therefore there is no manipulation. Manipulating span: No manipulation. Pristine span: No manipulation” \\
        \midrule
        LLM &  Opinion sentences: Ariana Grande’s New Lavender Hair Color Just Inspired Our Next Trip To The Salon”: Rationale: The tweet is only expressing opinion and therefore there is no manipulation. Manipulating span: No manipulation. Pristine span: No manipulation \\
        
        \midrule
        \multirow{2}{*}{User} &  Tweet: \textit{Input Tweet} \\
                              &   Article: \textit{Input Article}   \\
        \bottomrule
    \end{tabular}
    }
    
    \caption{The one-shot CoT prompt template to the LLMs. For two-shot and four-shot prompts, the prompt includes more examples. }
    \label{tab:llm_prompt}
    
\end{table*}

\label{apx:llm_prompts}

\section{Additional Qualitative Examples}
\label{apx:qualitative_examples}

\begin{table*}[t]
    \small
    \centering
    {
    \begin{tabular}{p{0.15\linewidth} p{0.8\linewidth}}
        \toprule
        
        \multicolumn{2}{c}{\textbf{Opinion Identification Error}} \\
        \midrule
        \multirow{2}{*}{Input} &  \textbf{Tweet}: Ariana Grande’s New Lavender Hair Color Just Inspired Our Next Trip To The Salon ... \\
                                &  \textbf{Article}: Ariana Grande Dyed Her Hair, And This Is Our Favorite Color Transformation Yet 
                                    Ariana Grande is giving us whiplash with her hairstyles lately, and we honestly love it. On July 18th, Grande took to Instagram to debut her latest hair transformation. She’s now sporting pastel lavender locks and good god (is a woman), it looks amazing on her...\\
        
        \arrayrulecolor{black!20}\midrule  
        
       \multirow{3}{*}{Prediction} & \textbf{Is manipulated}: Yes {\color{darkred} \xmark} \\
                                   & \textbf{Manipulating span}: Salon {\color{darkred} \xmark} \\
                                  & \textbf{Pristine span}: None\\
        \arrayrulecolor{black}\midrule 
        
        \multicolumn{2}{c}{\textbf{Pristine Span Extraction Failure}} \\
        \midrule
        \multirow{2}{*}{Input} &  \textbf{Tweet}: Transcript: Democratic Presidential Debate in \hlc{lightred}{Brooklyn} view more ... \\
                               &     \textbf{Article}: The Democratic Debate in \hlc{lightblue}{Cleveland} 
                                     This is rightly a big issue in Ohio. And I have laid out my criticism, but in addition my plan, for actually fixing NAFTA. Again, I have received a lot of incoming criticism from Senator Obama. And the Cleveland Plain Dealer examined Senator Obama's attacks on me regarding NAFTA and said they were erroneous. So I would hope that, again, we can get to a debate about what the real issues are and where we stand because we do need to fix NAFTA. It is not working. It was, unfortunately, heavily disadvantaging many of our industries, particularly manufacturing. ...\\
                                            
        \arrayrulecolor{black!20}\midrule  
        
       \multirow{3}{*}{Prediction} & \textbf{Is manipulated}: Yes \\
                                   & \textbf{Manipulating span}: Brooklyn\\
                                  & \textbf{Pristine span}: None {\color{darkred} \xmark}\\
        \arrayrulecolor{black}
        
        \bottomrule
    \end{tabular}
    }
    
    \caption{Example outputs from our baseline model where it produces erroneous outputs. } 
    
    \label{tab:qualitative}
\end{table*}

\begin{table*}[t]
    \small
    \centering
    {
    \begin{tabular}{p{0.15\linewidth} p{0.8\linewidth}}
        \toprule
        
        \multirow{2}{*}{Input} &  \textbf{Tweet}: Ariana Grande’s New Lavender Hair Color Just Inspired Our Next Trip To The Salon ...\\
                                &  \textbf{Article}: Ariana Grande Dyed Her Hair, And This Is Our Favorite Color Transformation Yet 
                                    Ariana Grande is giving us whiplash with her hairstyles lately, and we honestly love it. On July 18th, Grande took to Instagram to debut her latest hair transformation. She’s now sporting pastel lavender locks and good god (is a woman), it looks amazing on her...\\
        
        \arrayrulecolor{black!20}\midrule  
        
       \multirow{3}{*}{Prediction} & Is manipulated: Yes {\color{darkred} \xmark}  \\
                                   & Manipulating span: Salon {\color{darkred} \xmark}\\
                                  & Pristine span: None\\
        \arrayrulecolor{black}\midrule 

        \multirow{2}{*}{Input} &  \textbf{Tweet}: Ariana Grande’s New Lavender Hair Color Just Inspired Our Next Trip To The Salon ... \\
                                &  \textbf{Predicted Opinions}: Ariana Grande’s New Lavender Hair Color Just Inspired Our Next Trip To The Salon\\
                                &  \textbf{Article}: Ariana Grande Dyed Her Hair, And This Is Our Favorite Color Transformation Yet 
                                    Ariana Grande is giving us whiplash with her hairstyles lately, and we honestly love it. On July 18th, Grande took to Instagram to debut her latest hair transformation. She’s now sporting pastel lavender locks and good god (is a woman), it looks amazing on her...\\
        
        \arrayrulecolor{black!20}\midrule  
        
       \multirow{3}{*}{Prediction} & \textbf{Is manipulated}: No {\color{darkgreen} \cmark}\\
                                   & \textbf{Manipulating span}: None {\color{darkgreen} \cmark}\\
                                  & \textbf{Pristine span}: None\\

        \arrayrulecolor{black}
        
        \bottomrule
    \end{tabular}
    }
    
    \caption{Example outputs from our LED-FT and LLM + LED-FT. The predicted opinion extracted by ChatGPT allows the fine-tuned model to predict the manipulation label correctly. } 
    
    \label{tab:qualitative_correct}
\end{table*}

\Cref{tab:qualitative} presents two instances where our baseline model makes errors. In the first example, our model was not able to identify that ``Inspired Our Next Trip To The Salon'' is an expression of opinion, resulting in the model incorrectly classifying this sample as \recontextualized~. In the second example, although our model accurately predicts the example as \recontextualized~ and extracts the correct manipulating span, it fails to extract the pristine text span correctly, likely due to the nature of the training set, as discussed in \Cref{apx:remaining_challenges}.

\Cref{tab:qualitative_correct} shows an example where extracting opinion sentences from the tweet by ChatGPT enables our model to correctly identify the tweet as not manipulating the associated article. 

\section{Discourse Analysis Examples}
\label{apx:discourse_examples}
\Cref{tab:discourse_examples} shows examples of manipulated sentences associated with a \textit{Main} or \textit{Cause} discourse. A \textit{main} discourse implies that the sentence conveys the main story of an article, whereas a \textit{cause} discourse indicates that the sentences discuss the consequential aspect of the main story.

\begin{table*}[t]
    \small
    \centering
    {
    \begin{tabular}{p{0.15\linewidth} p{0.8\linewidth}}
        \toprule
        
        \multicolumn{2}{c}{\textbf{\textit{Main} Discourse}} \\
        \midrule
        Tweet &  \#Zuckerbergtestimony \hlc{lightred}{Mark Zuckerberg}'s testimony before the House Energy and Commerce Committee is over. \\
        \arrayrulecolor{black!20}\midrule
        Article                        & ... U.S. Rep. Joe Barton, R-Texas, chairman of the House Energy and Commerce Committee, made the following statement today during the full committee hearing on the Administration\'s FY \'07 Health Care Priorities: "Good afternoon.. \textbf{Let me begin by welcoming Secretary \hlc{lightblue}{Michael Leavitt} today to the Energy and Commerce Committee.} We look forward to hearing him testify about the Administration\'s Fiscal Year 2007 Health Care Priorities ...\\

        \arrayrulecolor{black}\midrule 
        
        \multicolumn{2}{c}{\textbf{\textit{Cause} Discourse}} \\
        \midrule
        Tweet &  Thank you, Rep. Johnson, for your service! Weekly Republican Address: Rep. \hlc{lightred}{Sam Johnson} (R-TX) ... via @YouTube  \\
        \arrayrulecolor{black!20}\midrule
        Article                       &      ... In the address, Boehner notes that this is a new approach that hasn’t been tried in Washington – by either party – and it is at the core of the Pledge to America, a governing agenda Republicans built by listening to the people. \textbf{Leader \hlc{lightblue}{Boehner} recorded the weekly address earlier this week from Ohio, where he ran a small business and saw first-hand how Washington can make it harder for employers and entrepreneurs to meet a payroll and create jobs.} Following is a transcript ...\\
                                            
        \arrayrulecolor{black}
        
        \bottomrule
    \end{tabular}
    }
    
    \caption{Examples of manipulated sentences with a \textit{Main} discourse and a \textit{Cause} discourse. The manipulated sentences are marked in \textbf{boldface}. The manipulating and pristine spans are marked in \hlc{lightred}{red} and \hlc{lightblue}{blue}, respectively.} 
    
    \label{tab:discourse_examples}
\end{table*}

\section{Error Type Definition}
\label{apx:error_type_defintion}
In this section, we provide illustrations for each error type:

\begin{itemize}
    \item \textbf{Opinion Identification Error}: The tweet predicts that a tweet manipulates the reference article. However, the manipulating span predicted by the model is in fact merely opinions and not trying to manipulate the content. An example is shown in Table 4 in the appendix. It is true that no annotator has specified the ground truth for opinion sentences. All the error analyses were performed manually by the authors.
    \item \textbf{Manipulation Detection False Alarm}: This is effectively the ``Manipulation Detection False Positive'' in which the model predicts a tweet manipulates the reference article but the label is \textsc{NoMani} (no manipulation). Note that ``Opinion Identification Error'' is considered a special case of ``Manipulation Detection False Alarm'' where the manipulating span overlaps with opinions expressed by the tweet author. 
    \item \textbf{Manipulation Detection False Negative}: The model predicts there is on manipulation within a tweet but the label is \textsc{Mani} (manipulating).
    \item \textbf{Manipulating Span Extraction Failure}: The model successfully predicts the manipulation label for a manipulating tweet but fails to identify the specific text spans that manipulate the content of the reference article.    
    \item \textbf{Pristine Span Extraction Failure}: The model successfully predicts the manipulation label for a manipulating tweet but fails to identify the specific text span from the reference article that was manipulated.
\end{itemize}

\end{document}